\documentclass[letterpaper, 10 pt, conference]{ieeeconf}
\IEEEoverridecommandlockouts
\usepackage{cite}
\usepackage{amsmath,amssymb,amsfonts}
\usepackage{algorithmicx}
\usepackage{graphicx}
\usepackage{textcomp}
\usepackage{xcolor}
\usepackage{algorithm}
\usepackage[noend]{algpseudocode}
\usepackage{subcaption}
\def\BibTeX{{\rm B\kern-.05em{\sc i\kern-.025em b}\kern-.08em
    T\kern-.1667em\lower.7ex\hbox{E}\kern-.125emX}}
    
\makeatletter
\newcommand\fs@spaceruled{\def\@fs@cfont{\bfseries}\let\@fs@capt\floatc@ruled
  \def\@fs@pre{\vspace{0.7\baselineskip}\hrule height.8pt depth0pt \kern2pt}%
  \def\@fs@post{\kern2pt\hrule\relax}%
  \def\@fs@mid{\kern2pt\hrule\kern2pt}%
  \let\@fs@iftopcapt\iftrue}
\makeatother
    
\newtheorem{problem}{Problem}

\newtheorem{remark}{Remark}

\newcommand{\X}{X}

\DeclareMathOperator*{\argmin}{argmin}

\begin{document}

\title{Anytime informed path re-planning and optimization for\\ robots in changing environments \\

\thanks{This work is partially supported by ShareWork project (H2020, European Commission -- G.A. 820807).}%
\thanks{$^{1}$ Dipartimento di Ingegneria Meccanica e Industriale, University of Brescia
    {\tt\small \{c.tonola001,manuel.beschi\}@unibs.it}}%
\thanks{$^{2}$ STIIMA-CNR - Institute of Intelligent Industrial Technologies and Systems, National Research Council of Italy
    {\tt\small \{marco.faroni,nicola.pedrocchi\}@stiima.cnr.it}}%
}

\author{
Cesare Tonola$^{1}$, Marco Faroni$^{2}$, Nicola Pedrocchi$^{2}$, Manuel Beschi$^{1,2}$
}%

\maketitle

\begin{abstract}
In this paper, we propose a path re-planning algorithm that makes robots able to work in scenarios with moving obstacles.
The algorithm switches between a set of pre-computed paths to avoid collisions with moving obstacles.
It also improves the current path in an anytime fashion.
The use of informed sampling enhances the search speed.
Numerical results show the effectiveness of the strategy in different simulation scenarios.
\end{abstract}


\section{Introduction}
Path planning is important in many fields, such as robotics, computer science, aerospace, and aeronautics. 
It deals with finding a path (\emph{i.e.}, a sequence of states) from a start position to a goal position. 
In robotics, a feasible solution is a collision-free path that satisfies the system dynamics constraints. 
Many techniques have been proposed to solve this problem. 
Graph-based searches, such as A* \cite{A_star}, offer properties of completeness and optimality but they suffer from the curse of dimensionality.
This issue is mitigated in sampling based methods, such as RRT \cite{RRT}, by randomly sampling the search space.
For this reason, sampling-based algorithms are the most widespread when it comes to high-dimensional systems such as robot manipulators. 
While in the past the feasibility of the solution was the main concern, recent methods have also addressed the problem of finding an optimal solution with respect to a given objective \cite{karaman:RRT*} \cite{Gammell2020}.

Many robotic applications have a limited planning time to find a solution and speeding up the convergence rate of optimal planners is thus a relevant field of research.
This occurs, for example, when the robot operates in dynamic environments.
Recently, Gammel et al. proposed the concept of informed sampling \cite{Gammel:InformedRRT}; that is, shrinking the sampling space to an hyper-ellipsoid that contains nodes with non-null probability to improve the current solution.
Another strategy to tackle the limited computing time is the so-called \emph{anytime search} \cite{Gammell:BIT} \cite{Aine2016}. 
In practice, a first sub-optimal solution is found in a short time and the robot starts executing it.
Then, the solution is improved iteratively during its execution.
Path planning in robotics should also deal with moving obstacles, moving goals, and unstructured environments.
For this reason, online re-planning is essentials to operate in a real world. 
For example, it is gaining more and more importance in the Human-robot collaboration (HRC). 
Nowadays, robots are enclosed into cells or they stop or reduce their speed when an operator approaches \cite{Magrini:coexistance-interaction}.
Real-time techniques exist that reduce safety stops and optimize the  speed reduction on a pre-defined path, for example via linear programming \cite{Zanchettin:safety_HRC}, PID control \cite{Faroni_UR2020}, or model predictive control \cite{Faroni_ETFA2019}.
However, in order for robots to react more naturally to the operator's movements, they should also learn how to rapidly change their path when humans interfere with their motion.
In this context, path re-planning plays a key role.

Over the years some strategies have been implemented to calculate online new paths or to modify the current one. 
Some methods are variants of the well-known RRT and they try to reuse historical information about the state space. 
Some of them are designed for multiple-query plannings problem like RRF \cite{RRF}, that builds a forest of disconnected RRTs rooted at different locations which try to connect to each others. 
Other algorithms are suitable for single-query planning problems. DRRT \cite{DRRT} and \cite{Connell:DRRT*} regrow a new tree trying to reuse the still-valid portion of the previous RRT tree when a new obstacle appears. MP-RRT \cite{MP-RRT} combines the concept of RRF forests, tree-reuse of DRRT and waypoints cache of ERRT \cite{ERRT}. RRT\textsuperscript{x} \cite{RRTX} repairs the same search tree over the entire navigation rather than growing a new one.
All these techniques prune trees when changes of the configuration space happen. 
However, when the environment is complex, a larger effort is required to prune the graph rather than re-plan a new one \cite{HL-RRT*}.
In some approaches, time dimension is added to the tree to plan a new path foreseeing possible future collisions with mobile obstacle \cite{Zhang2021}. 
Obstacles as time-space volumes can be used to check vertices and edges inside a defined time horizon, postponing the check of those outside it \cite{van-den-Berg:anytime,HL-RRT*}.
Another type of strategy involves the use of potential fields in the configuration space. 
A free trajectory is calculated following the negative gradient of the potential. 
For example, in \cite{elastic-strips}, an initial trajectory is deformed by a force dependent on the distance between the robot and the obstacle; then, another force tends to restore the trajectory to its initial structure. 
This method can suffer from local minima, furthermore it modifies a trajectory but remains tied to it. If the environmental change results in a passage elimination, a solution may not be found (\emph{i.e.}, it is not complete).

\subsection{Contribution}
Online path re-planning and path optimization are two main topics in motion planning. 
In this paper, we propose an online re-planning framework capable of both tasks.
The rationale behind this strategy is that using a set of pre-computed paths to re-plan the current one reduces the computational load and allows the exploration of solutions completely different from each other. 
Furthermore, the cost of the best solution found so far is taken into account to avoid the search for solutions that are not able to improve the current path.
The main contributions of the proposed framework are:
\begin{itemize}
    \item It combines re-planning and path-optimization strategies in an anytime fashion to improve the solution over time;
    \item It proposes two algorithms, \emph{informedOnlineReplanning} and \emph{pathSwitch}, to both re-plan and improve the current path connecting it to the paths of the set;
    \item It uses informed sampling based on the best solution found so far to enhance the optimality convergence rate;
    \item The number of collision checks is reduced because it is a path-based and not a tree-based approach;
    \item It does not suffer from local minima and it does not have a strong dependence on the path initially followed.
    \item It is independent from the sampling-based algorithm used for the search of the paths;
\end{itemize}

The paper is structured as follows: problem formulation is given in Section \ref{sec:problem-formulation}, the overall framework is described in Section \ref{sec:strategy}. 
Numerical experiments are provided and discussed in Section \ref{sec:result}.
Finally, conclusions are drawn in Section \ref{sec:conclusions}.

\section{Problem formulation}
\label{sec:problem-formulation}

Path planning finds a collision-free path from a given start position to a desired goal position (or set of positions).
The problem is formulated in the configuration space $\X$, defined by all possible robot configurations $x$.
For robot manipulators, $x$ is usually a real vector of joint positions.
Let $\X_{\mathrm{obs}} \subset \X$ be the space of all points in collision with obstacles; then the search space of the problem is given by the free space $\X_{\mathrm{free}} = \mathrm{cl}(\X \setminus \X_{\mathrm{obs}}$), where $\mathrm{cl}(\cdot)$ is the closure of a set.
Therefore, we consider the following path planning problem.
\begin{problem}
    Given an initial configuration $x_{\mathrm{start}} \in \X$ and a goal configuration $x_{\mathrm{goal}} \in \X$, a path planning problem finds a curve $\sigma:[0,1] \rightarrow   \X_{\mathrm{free}}$ such that $\sigma(0)=x_{\mathrm{start}}$ and $\sigma(1) = x_{\mathrm{goal}}$.
    A solution curve to such a problem is a \emph{feasible path}.
\end{problem}

One may seek for the feasible path that optimizes a given objective.
To this purpose, consider a cost function $c: \Sigma_{\mathrm{free}} \rightarrow \mathbb R$ that associates a cost with any feasible path $\sigma \in \Sigma_{\mathrm{free}}$.
An \emph{optimal path} is a feasible path $\sigma^*$ such that:
\begin{equation}
\label{eq:optimal-motion-planning}
    \sigma^*=\argmin_{\sigma\in\Sigma_{\mathrm{free}}} c(\sigma)
\end{equation}
The cost function $c$ is the length of the path, denoted by $ \left\Vert \sigma \right \Vert$, so that the optimal motion plan is the shortest collision-free path from $x_{\mathrm{start}}$ to $x_{\mathrm{goal}}$.

Most times, path planning problems assume that $X_{\mathrm{obs}}$ is constant over time.
Such an assumption is reasonable if the environment is structured (\emph{e.g.}, a fenced robotized cell), but it does not hold if the environment is unstructured.
In this work, we consider the case in which $\X_{\mathrm{obs}}$ changes unpredictably over time and it is necessary to deploy a reactive behavior that modifies the robot motion at runtime, to avoid collisions and reach the desired goal.

\emph{Online path re-planning} implements such a reactive behavior by modifying an initial path during its execution.
Namely, the robot starts executing a feasible path and, as soon as that path becomes infeasible because of a moving obstacle, it seeks for a new path from its current state to the goal.

\subsection{Notation}
\label{subsec:notation}

The following notation is adopted throughout the paper:
\begin{itemize}
    \item $\sigma_i : [0,1] \rightarrow \X_{\mathrm{free}}$ is a path from $x_{\mathrm{start}}$ to $x_{\mathrm{goal}}$;
    \item $s_i = (x_1,\dots,x_M)$ is the sequence of waypoints (nodes) that composes $\sigma_i$, where $x_1 = x_{\mathrm{start}}$ and $x_M = x_{\mathrm{goal}}$;
    \item $\sigma_i[x_j,x_k]$ is the portion of $\sigma_i$ from a $x_j \in \sigma_i$ to a point $x_k \in \sigma_i$;
    \item $s_i[j,k] = (x_j,\dots,x_k)$, where $j\geq 1$ and $k\leq M$ is a sub-sequence of $s_i$. Similarly, $s_i[j] = x_j$;
    \item $c: \Sigma \rightarrow \mathbb R_{\geq 0}$ is a cost function that associates a positive real cost to a feasible path and $+\infty$ if the path is infeasible;
    \item $\left\Vert \sigma \right\Vert :  \Sigma \rightarrow \mathbb R_{\geq 0}$ is the length of a path $\sigma \in \Sigma$;
    \item $\sigma_i \cup \sigma_j: \Sigma \rightarrow \Sigma$ is a function that concatenates $\sigma_j$ with $\sigma_i$ (being $\sigma_i(1)=\sigma_j(0)$).
\end{itemize}

\begin{remark}
\label{remark2}
	Note that, with respect to  (\ref{eq:optimal-motion-planning}), the domain of $c$ is extended to the set of all paths $\Sigma$ by assigning an infinite cost with any infeasible paths (\emph{e.g.} paths obstructed by an obstacle).
\end{remark}

\section{Path re-planning strategy}
\label{sec:strategy}
\subsection{Re-planning scheme}
\label{subsec:scheme}
The re-planning strategy presented in this paper has a double functionality. The algorithm is able to compute a new free path when the current one becomes infeasible, but  it is also able to optimize the current path during its execution. 
The proposed re-planning scheme consists of three threads running in parallel, as shown in Algorithm \ref{alg:threads}:
\begin{itemize}
    \item The \emph{trajectory execution thread}: it receives a set of feasible paths $S$ and the path $\sigma_i \in S$ to be executed. It sends the corresponding joint commands to the the robot controller at a high rate.
    \item The \emph{collision checking thread}: it verifies whether each path $\sigma_j \in T$ is in collision or not during the execution of the trajectory. $T$ is derived from $S$ replacing the current path $\sigma_i \in S$ with $\sigma_i [x_h,x_{\mathrm{goal}}]$, which is the part of $\sigma_i$ from the robot configuration $x_h$ to the goal. For each path $\sigma_j \in T$, it computes a boolean variable equal to $0$ if $\sigma_j$ is collision free and equal to $1$ otherwise. Moreover, it computes the nodes right before and after the obstacles, $x_{\mathrm{before}} \in  s_j$ and $x_{\mathrm{after}} \in  s_j$ respectively (see Figure \ref{pathSwitch}).
    \item The \emph{path re-planning thread}: it invokes the re-planning algorithm to find a feasible solution when an obstacle is obstructing the current path or to optimize it. 
    When a new path is found, the trajectory is computed and it is executed by the \emph{trajectory execution thread}.
\end{itemize}

The path re-planning thread exploits two algorithms which communicate with each other. The first one, \emph{pathSwitch}, searches for a path that starts from a given node of the path currently traveled by the robot, towards each of the other available paths. The second one, \emph{informedOnlineReplanning}, manages the whole re-planning procedure: it feeds \emph{pathSwitch} with a set of available paths and it defines a set of nodes from which starting \emph{pathSwitch}. 
This strategy is based on an anytime approach, so the aim is to get a first feasible solution in a very short time and then try to improve such a path during execution. The next two subsections describe \emph{pathSwitch} and \emph{informedOnlineReplanning} in detail.

\renewcommand{\algorithmicrequire}{\textbf{Input:}}
\algrenewcommand\algorithmicloop{\textbf{Thread}}
\floatstyle{spaceruled}
\restylefloat{algorithm}
\begin{algorithm}[tpb]
\caption{Threads operating in parallel}
\label{alg:threads}
\small
\begin{algorithmic}[1]
\Require{set of paths $S=\{\sigma_1,\dots,\sigma_N\}$, index $i$ of the current path}
\Loop\; \emph{Re-planning}
\State $x_h \leftarrow \mathrm{projectOnPath}(\mathrm{state},\sigma_i)$\;
\State $\sigma_{\mathrm{RP}} \leftarrow  \mathrm{informedOnlineReplanning}(S,i,x_h)$\;
\State $\mathrm{trj} \leftarrow \mathrm{computeTrajectory}(\sigma_{\mathrm{RP}})$ \;
\State $\sigma_i \leftarrow \sigma_{\mathrm{RP}}$\;
\EndLoop
\vspace{-0.1cm}
\Statex \hrulefill \vspace{0.1cm}
\Loop\; \emph{Collision check} \;
\State $T \leftarrow S \setminus \{\sigma_i\}$\;
\State $T \xleftarrow{+} \sigma_i[x_h,x_{\mathrm{goal}}]$
\For {${\sigma_j \in T}$}
    \State $(\mathrm{coll\_res}, x_{\mathrm{before}}, x_{\mathrm{after}}) \leftarrow \mathrm{checkCollision}(\sigma_j)$\;
    \If {$\mathrm{coll\_res}$ is $\mathrm{True}$}
        \State $c_{\sigma_j} \leftarrow +\infty$\;
        \If {$\sigma_j = \sigma_i[x_h,x_{\mathrm{goal}}]$}
            \State $t_{\mathrm{RP}} \leftarrow \mathrm{reducedTime}$ \;
        \EndIf
    \Else
        \State $c_{\sigma_j} \leftarrow c(\sigma_j)$\;
        \If {$\sigma_j = \sigma_i[x_h,x_{\mathrm{goal}}]$}
            \State $t_{\mathrm{RP}} \leftarrow \mathrm{relaxedTime}$ \;
        \EndIf
    \EndIf
\EndFor

\EndLoop
\vspace{-0.1cm}
\Statex \hrulefill \vspace{0.1cm}
\Loop\; \emph{Trajectory execution} \;
\State $\mathrm{state} \leftarrow \mathrm{sampleTrajectory}(\mathrm{trj},t)$ \;
\State $\mathrm{sendToController}(\mathrm{state})$ \;
\EndLoop

\end{algorithmic}
\normalsize
\end{algorithm}

\subsection{PathSwitch algorithm}
\label{subsec:pathswitch}
\emph{pathSwitch} aims to create a path from a node of the path currently traveled to each node of a given set of paths $P$.
The procedure is described in Algorithm \ref{alg:pathswitch}.
The inputs of the algorithm are the starting node $x_n$, a set of available paths $P=\{\sigma_1,\dots,\sigma_N\}$, the current path $\sigma_i \notin P$, and the maximum allowed computing time $t_{\mathrm{max}}$.
The output is the best path $\sigma_{\mathrm{switch}}$ from $x_n$ to $x_{\mathrm{goal}}$ found so far.

For all paths $\sigma_j \in P$, the algorithm searches for a path from $x_n$ to the nodes of $\sigma_j$ (ordered based on the distance from $x_n$).
The search is performed by means of a sampling-based path planner in function \emph{planInEllipsoid}.
The result is the path $\sigma_{\mathrm{conn}}$.
If the path from $x_n$ to $x_{\mathrm{goal}}$, given by the concatenation of $\sigma_{\mathrm{conn}}$ and $\sigma_j[x_j,x_{\mathrm{goal}}]$, is better than the current one, it is stored as the best solution so far.
The procedure is interrupted when all paths in $P$ have been evaluated or when the computing time exceeds the maximum allowed time $t_{\mathrm{max}}$.

\begin{figure}[tbp]
\centerline{\includegraphics[width=0.5\textwidth]{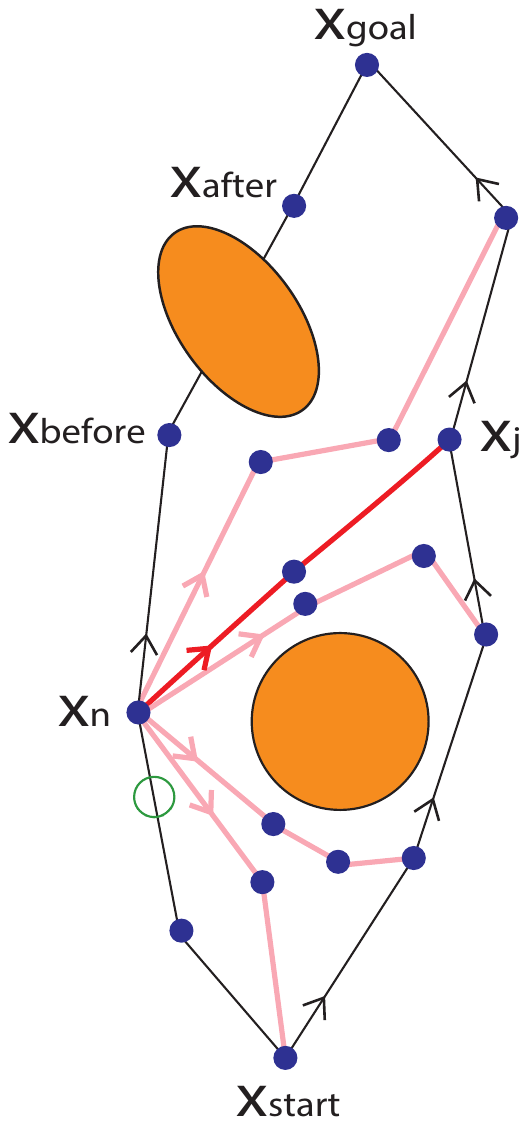}}
\caption{Example of how \emph{pathSwitch} works. Green circle: robot current configuration; orange shapes: moving obstacles. The algorithm searches for the connecting paths from $x_n \in \sigma_i$  to each node of $\sigma_j$ (pink lines) and selects the one that minimizes the overall cost from $x_n$ to $x_{\mathrm{goal}}$ (red lines).}
\label{pathSwitch}
\end{figure}

Figure \ref{pathSwitch} is an example of how the algorithm works: the green circle represents the current robot configuration and the orange shapes are two obstacles at the current time. 
\emph{pathSwitch} searches for the connecting paths from $x_n \in \sigma_i$  to each node of $\sigma_j$ (pink paths) and selects the one that minimizes the overall cost from $x_n$ to $x_{\mathrm{goal}}$ (red path).

\subsection*{1) Heuristics for a faster search of connecting paths} \label{conditions}

We speed up the search performed in \emph{pathSwitch} by using two strategies: i) excluding connecting nodes that can not improve the current solution; ii) using informed planning to reduce the search space of a connecting path.

\begin{figure}[tpb]
\centerline{\includegraphics[width=0.45\textwidth]{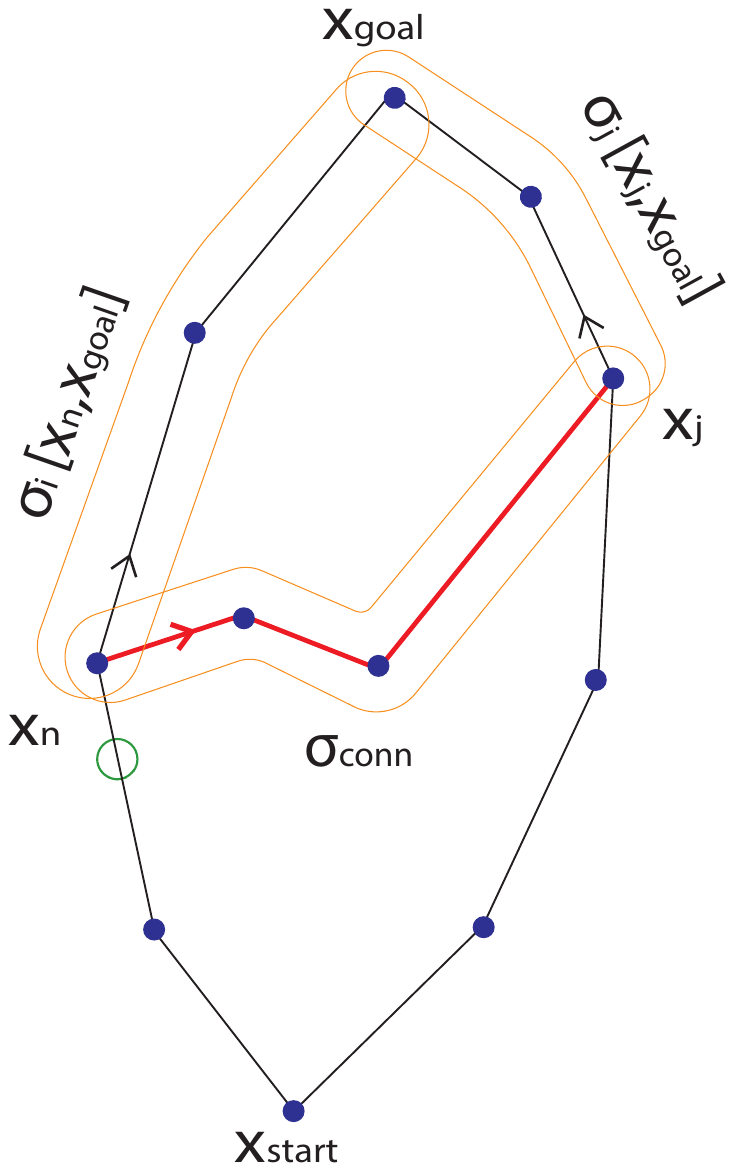}}
\caption{Representation of the nodes and subpaths considered to reduce the computational load of the algorithm.}
\label{computational_load}
\end{figure}

Referring to Figure \ref{computational_load}, let $x_n \in \sigma_i$ be the root node of \emph{pathSwitch} and $x_j \in \sigma_j$ the goal node of the connecting path $\sigma_{\mathrm{conn}}$.
In order for the candidate solution $\sigma_{\mathrm{switch}} = \sigma_{\mathrm{conn}} \cup \sigma_j[x_j,x_{\mathrm{goal}}]$ to be better than $\sigma_i$[$x_n$,$x_{\mathrm{goal}}$], the following condition must hold:
\begin{equation}
c(\sigma_{\mathrm{conn}}) <  c(\sigma_i[x_n,x_{\mathrm{goal}}]) -c(\sigma_j[x_j,x_{\mathrm{goal}}])
\label{eq1}
\end{equation}
The lower bound to the path length of $\sigma_{\mathrm{conn}}$ is the Euclidean distance from $x_n$ to $x_j$.
Consequently, a necessary condition for $x_j$ to improve the current path is that:
\begin{equation}
\left\Vert x_n - x_j \right\Vert <  c(\sigma_i[x_n,x_{\mathrm{goal}}]) -c(\sigma_j[x_j,x_{\mathrm{goal}}])
\label{eq2}
\end{equation}
If \eqref{eq2} does not hold, \emph{pathSwitch} skips node $x_j$ (lines 6-7 of Algorithm \ref{alg:pathswitch}).
Note that, according to \emph{Remark \ref{remark2}}, if $\sigma_i[x_n,x_{\mathrm{goal}}]$ is infeasible, \eqref{eq1} and \eqref{eq2} always hold. 

When a first solution has been found, \eqref{eq2} is updated with the cost of the new path $c(\sigma_{\mathrm{switch}})$, resulting in:
\begin{equation}
\left\Vert x_n - x_j \right\Vert < c(\sigma_{\mathrm{switch}}) - c(\sigma_j[x_j,x_{\mathrm{goal}}])
\label{eq3}
\end{equation}
In this way, only nodes that can improve the final solution are taken into consideration. Starting with the closer node as the first node to connect to is a good way to prune the calculations. 
Furthermore, when the distance between successive nodes of $\sigma_j$ is less than a certain threshold, only one of them is considered, because they would not bring very different solutions from each other.

As a second strategy to enhance the speed of \emph{pathSwitch}, the function \emph{planInEllipsoid} (line 8 of Algorithm \ref{alg:pathswitch}) relies on informed sampling \cite{Gammel:InformedRRT}.
In brief, when searching for a connecting path from $x_n$ to $x_j$, it is possible to shrink the sampling space to the following hyper-ellipsoid:
\begin{equation}
X_{\mathrm{ell}} = \{x \in X_{\mathrm{free}} \mid \Vert x-x_n \Vert + \Vert x_j - x \Vert < c_i \}
\label{eq4}
\end{equation}
where:
\begin{equation}
c_i = c(\sigma_{\mathrm{switch}}) - c(\sigma_j[x_j,x_{\mathrm{goal}}])
\label{eq5}
\end{equation}
Being \eqref{eq4} an admissible heuristic set, the nodes outside the ellipsoid can not improve the current solution and they can be discarded.

\renewcommand{\algorithmicrequire}{\textbf{Input:}}
\renewcommand{\algorithmicensure}{\textbf{Output:}}
\floatstyle{spaceruled}
\restylefloat{algorithm}
\begin{algorithm}[tpb]
\caption{PathSwitch algorithm}
\label{alg:pathswitch}
\small
\begin{algorithmic}[1]
\Require{node $x_n \in \sigma_i$, the current path $\sigma_i \notin P$, set of paths $P=\{\sigma_1,\dots,\sigma_N\}$, max time $t_{\mathrm{max}}$}
\Ensure{$\sigma_{\mathrm{switch}}$}

\State $\sigma_{\mathrm{switch}}$ $\leftarrow \sigma_i[x_n,x_{\mathrm{goal}}] $\;
\For {$\sigma_j \in P$}
    \State $Q \leftarrow s_{j}$
    \While {$\neg\,$ $\mathrm{isEmpty}(Q)$ \& $\neg\,$ $\mathrm{timeExpired}(t_{\mathrm{max}})$}
            \State $x_j \leftarrow \argmin_{x \in Q} || x-x_n|| $\;\;
            \State $\mathrm{max\_cost}$ $\gets$ $c(\sigma_{\mathrm{switch}}) - c(\sigma_j[x_j,x_{\mathrm{goal}}])$\;
            \If {$\left\Vert x_n - x_j \right\Vert < \mathrm{max\_cost}$}
                \State $\sigma_{\mathrm{conn}} \leftarrow \mathrm{planInEllipsoid}(x_n,x_j)$\;
                \If {$\neg\,$ $\mathrm{isEmpty}(\sigma_{\mathrm{conn}})$}
                    \If {$c(\sigma_{\mathrm{conn}}) + c(\sigma_j[x_j,x_{\mathrm{goal}}]) < c(\sigma_{\mathrm{switch}})$}
                        \State $\sigma_{\mathrm{switch}}$ $\gets$  $\sigma_{\mathrm{conn}} \cup $ $\sigma_j[x_j,x_{\mathrm{goal}}]$\;
                    \EndIf
                \EndIf
            \EndIf
            \State $Q.\mathrm{remove}(x_j)$\;
    \EndWhile
\EndFor
\end{algorithmic}
\normalsize
\end{algorithm}

\subsection{InformedOnlineReplanning algorithm}
\label{subsec:informedonlinereplanning}
\emph{informedOnlineReplanning} (Algorithm \ref{alg:informedonlinereplanning}) manages the whole re-planning procedure, by calling several times \emph{pathSwitch}, and giving the required inputs to it. 
\emph{informedOnlineReplanning} calls \emph{pathSwitch} giving it a different starting node and the updated set of available paths $P$. 
The set $P$ is obtained from the set $S$ of available paths calculated before starting the movement and replacing the current path $\sigma_i \in S$  with $\sigma_i[x_{\mathrm{after}},x_{\mathrm{goal}}]$, which is the part of the current path that lies beyond the obstacle. 
The set of nodes is determined by the mutual position between the obstacle and the configuration of the robot. 
If the obstacle obstructs the connection on which the robot configuration resides, the re-planning must start from the configuration itself. 
Otherwise, there are multiple free nodes from which \emph{pathSwitch} can be called. 
The idea is to start from the farthest node from the current configuration to have enough time to find a new solution before traveling through the node.
When \emph{pathSwitch} is called from a node, the cost of the best solution found up to that moment is used to search for better and better solutions. 
Furthermore, when all the nodes have been used and a solution has been found, the nodes of the candidate solution that have not been evaluated are added to the set. 
The algorithm ends when all the available nodes have been analyzed or when the computing time exceeds the maximum allowed time $t_{\mathrm{RP}}$, as explained in Section \ref{subsec:time}. 
The procedure is repeated in loop, as shown in Algorithm \ref{alg:threads}.
Note that, if \emph{informedOnlineReplanning} fails to find a solution when the current path is infeasible,  a contingency plan should be implemented in the \emph{trajectory execution thread} to avoid collisions (\emph{e.g.}, a safety stop should be issued).

\renewcommand{\algorithmicrequire}{\textbf{Input:}}
\renewcommand{\algorithmicensure}{\textbf{Output:}}
\algrenewcommand\algorithmicloop{\textbf{Thread}}
\floatstyle{spaceruled}
\restylefloat{algorithm}
\begin{algorithm}[tpb]
\caption{InformedOnlineReplanning algorithm}
\label{alg:informedonlinereplanning}
\small
\begin{algorithmic}[1]
\Require{set of paths $S=\{\sigma_1,\dots,\sigma_N\}$, index $i$ of the current path, current robot configuration $x_h \in \sigma_i$}
\Ensure{re-planned path $\sigma_{\mathrm{RP}}$}
\State $\sigma_{\mathrm{cur}} = \sigma_i [x_{\mathrm{h}},x_{\mathrm{goal}}]$;
\State $P \leftarrow S \setminus \{\sigma_i\}$\;
\State \footnotesize $(c_{\mathrm{cur}}, x_{\mathrm{before}}, x_{\mathrm{after}}, t_{\mathrm{RP}}) \leftarrow \mathrm{getFromCollisionThread}(\sigma_{\mathrm{cur}})$\;
\small
\If {$c_{\mathrm{cur}}$ $=$ $\infty$}
    \State $P \xleftarrow{+} \sigma_{\mathrm{cur}}[x_{\mathrm{after}},x_{\mathrm{goal}}]$\;
    \State $Q \leftarrow s_{\mathrm{cur}}[x_h,x_{\mathrm{before}}]$\;
\Else
    \State $P \xleftarrow{+} \sigma_{\mathrm{cur}}$\;
    \State $Q \leftarrow s_{\mathrm{cur}}$\;
\EndIf
\State $\sigma_{\mathrm{RP}} \leftarrow \sigma_{\mathrm{cur}}$\;
\While {$\neg\,$ $\mathrm{isEmpty}(Q)$ \& $t < t_{\mathrm{RP}}$}
    \State $x_n \leftarrow \argmin_{x \in Q} || x-x_{\mathrm{goal}}|| $\;
    \State $t_{\mathrm{max}} \leftarrow t_{\mathrm{RP}} - t$ \;
    \State $\sigma_{\mathrm{switch}} \leftarrow$ $\mathrm{pathSwitch}(x_n,\sigma_{\mathrm{RP}},P,t_{\mathrm{max}})$\;
    \If{$\neg\,$ $\mathrm{isEmpty}(\sigma_{\mathrm{switch}})$}
    \State $\sigma_{\mathrm{new}} \leftarrow \sigma_{\mathrm{cur}}[x_h,x_n]  \cup \sigma_{\mathrm{switch}}$ \;
        \If {$c(\sigma_{\mathrm{new}}) < c_{\mathrm{cur}}$}
            \State $\sigma_{\mathrm{RP}}\leftarrow \sigma_{\mathrm{new}}$ \;
            \State $c_{\mathrm{cur}} \leftarrow c(\sigma_{\mathrm{new}})$\;
        \EndIf
    \EndIf
    \State $Q.\mathrm{remove}(x_n)$\;
    \If{$\mathrm{isEmpty}(Q)$ \& $\sigma_{\mathrm{RP}} \not= \sigma_{\mathrm{cur}}$}
    \For{$x_{RP} \in s_{\mathrm{RP}}$}
        \If{$\neg\,$ $\mathrm{alreadyUsed}(x_{RP})$}
            \State $Q.\mathrm{add}(x_{RP})$\;
        \EndIf
    \EndFor
    \EndIf
\EndWhile
\end{algorithmic}
\normalsize
\end{algorithm}

\subsection{Time constraints}
\label{subsec:time}

The re-planning algorithm is executed in loop, with a maximum allowed cycle time $t_{\mathrm{RP}}$.
The value of $t_{\mathrm{RP}}$ depends on whether the current path is deemed to be feasible or not by the \emph{collision checking thread}.
When the current path is infeasible, a new path must be found as fast as possible. 
In this case, a short time is given to the algorithm to quickly obtain a feasible trajectory for the robot; the priority is finding a solution rather than improving its cost. 
Otherwise, when the current path is feasible, the aim is to improve the path reducing its cost and $t_{\mathrm{RP}}$ is larger so the algorithm can conduct a deeper search towards better solutions.
The value of $t_{\mathrm{RP}}$ is set by the \emph{collision checking thread} (Algorithm \ref{alg:threads}, lines 14 and 18).

Let \emph{pathSwitch cycle} be the iteration during which \emph{pathSwitch} tries to find a path starting from the given node $x_n \in s_i$ to a selected node $x_j \in s_j$ of an available path. When the current path is obstructed, the whole remaining available time $t_{\mathrm{max}}$ is given to the \emph{pathSwitch cycle}. Until a feasible solution is found the only time constraint is to not exceed $t_{\mathrm{max}}$. Then, when a path that avoids the obstacle has been found, the new priority is to improve the solution found. From this moment, the next \emph{pathSwitch cycles} of the same call to \emph{pathSwitch} are required to not use a time greater than the average of the time required by the previous successful cycles. This is done to not spend the whole remaining time trying to connect $x_n \in s_i$ to a node $x_j \in s_j$ particularly difficult to be reached due to interposed obstacles. At the end of a cycle, if the remaining time is less than the previous mentioned time average, a new cycle will not start and the call to \emph{pathSwitch} ends. Similarly, let's call \emph{informedOnlineReplanning cycle} the iteration during which \emph{informedOnlineReplanning} calls \emph{pathSwitch} from a given node  $x_n \in s_i$. If the elapsed time at the end of the cycle exceeds the maximum time $t_{\mathrm{RP}}$, the algorithm stops. \emph{timeExpired}() at line 4 of Algorithm \ref{alg:pathswitch} and line 11 of Algorithm \ref{alg:informedonlinereplanning} verify these conditions.

\section{Simulations and results}
\label{sec:result}
The proposed framework has been simulated using ROS and \emph{MoveIt!} on a laptop with a $2.80$~GHz 8-core CPU. The method has been tested in two different scenarios:
\begin{itemize}
    \item A point robot moving in a 3D space, where a large obstacle composed of four overlapped boxes is placed between the robot initial position and the goal configuration;
    \item A robotized cell with a 6-degree-of-freedom anthropomorphic robot and a cylindrical fixed obstacle placed between the robot initial position and the goal configuration.
\end{itemize}
The initial set of paths $S$ is computed using RRT-Connect \cite{Lavalle:RRTconnect} solver and then they are optimized with RRT* \cite{karaman:RRT*}.
The \emph{trajectory execution thread} runs at $100$~Hz and the \emph{collision checking thread} runs at $30$~Hz. 
The frequency of the \emph{re-planning thread} depends on the time given to \emph{informedOnlineReplanning}, as explained in Section \ref{subsec:time}.
According to the naming given in Algorithm \ref{alg:threads}, we set $\mathrm{reducedTime}$ = $50$ ms and $\mathrm{relaxedTime}$ = $100$ ms in the 3D scenario; and $\mathrm{reducedTime}$ = $70$ ms and $\mathrm{relaxedTime}$ = $120$ ms  in the 6D scenario.

The tests consist of 30 iterations in which the following steps are executed:
\begin{itemize}
    \item four paths are computed from the start to the goal; the start and goal configurations have been chosen so as to be separated by the fixed obstacle; one of this path is the current path of the robot, the others compose the set of available paths for the re-planner;
    \item the robot starts following the current path and at time instants $0.5$, $1.0$ and $1.5$~s three cubic obstacles with side of $0.05$~m obstruct a random connection of the path from the current robot configuration to the goal; one of them always obstructs the connection crossed by the robot configuration at that time, to add further complexity to the re-planning;
    \item every time the re-planner finds a solution, the robot starts following it. The re-planning time, the length of the previous (possibly infeasible) path from the configuration to the goal and the length of the solution just found are saved.
\end{itemize}

\begin{figure*}
\vspace{0.3cm}
\centering
    \subcaptionbox{The green path is the current path $\sigma_i$, the yellow one is the result of the re-planner path optimization $\sigma_{\mathrm{RP}}$, the others are the available paths $\sigma_j \in S$.\label{subfig:init}}
    {\includegraphics[width=0.4\linewidth]{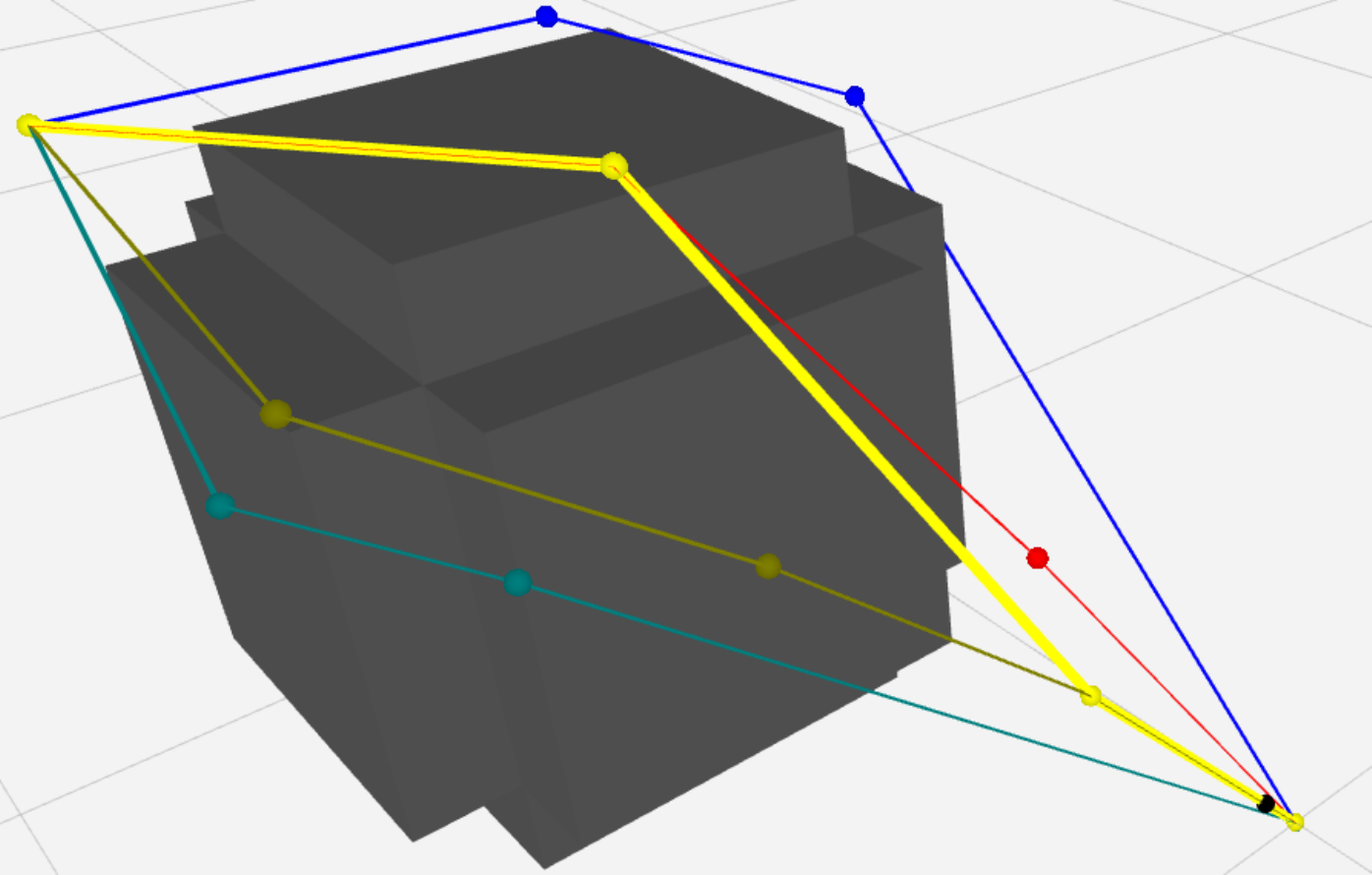}}\hfil
    \subcaptionbox{An obstacle obstructs the path $\sigma_{\mathrm{RP}}$. \label{subfig:obstacle}}
    {\includegraphics[width=0.4\linewidth]{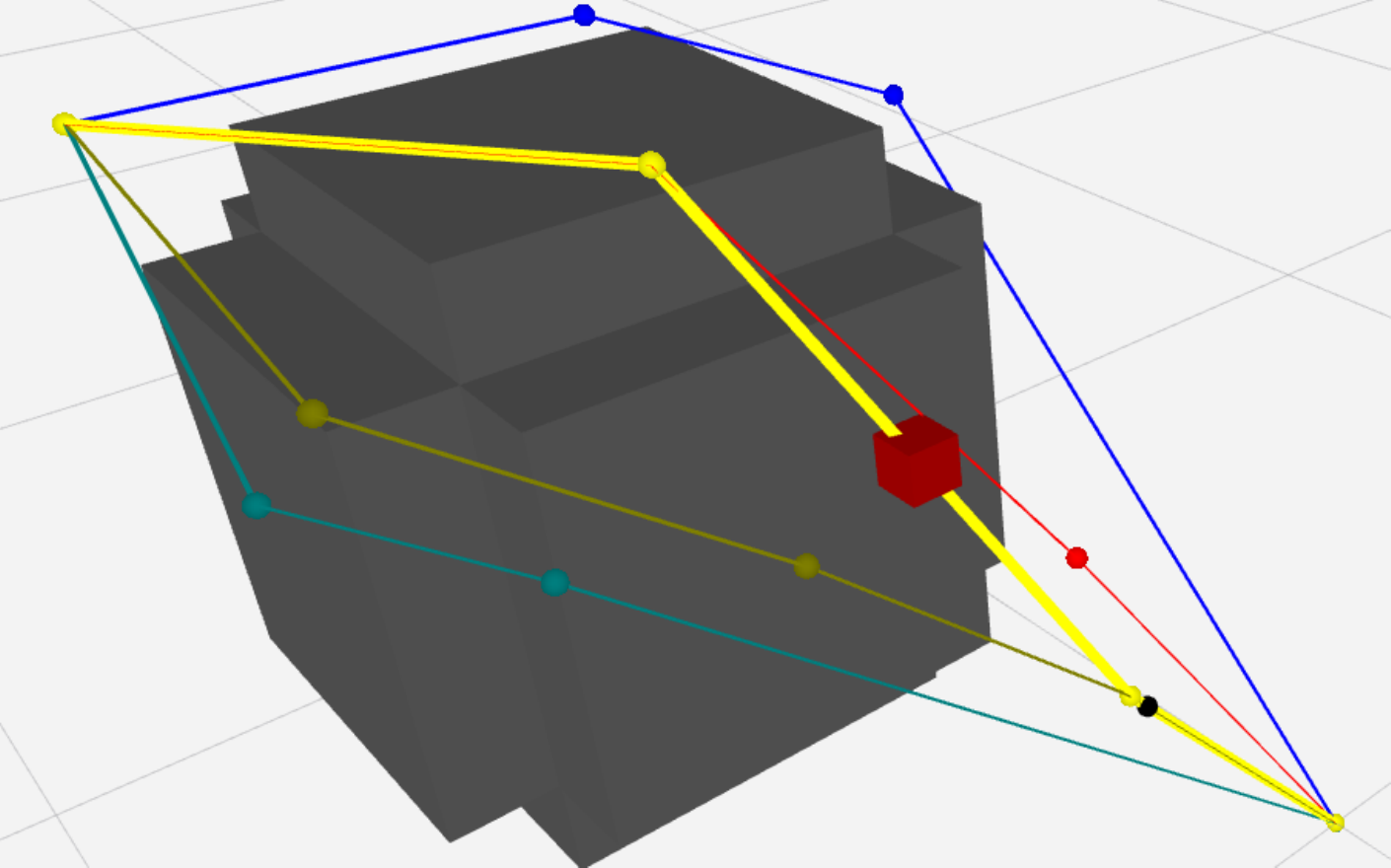}}\par\medskip
    \subcaptionbox{The yellow path is the first feasible path found after the obstacle appearance, it will be optimized in the next iterations. \label{subfig:avoidance}}
    {\includegraphics[width=0.4\linewidth]{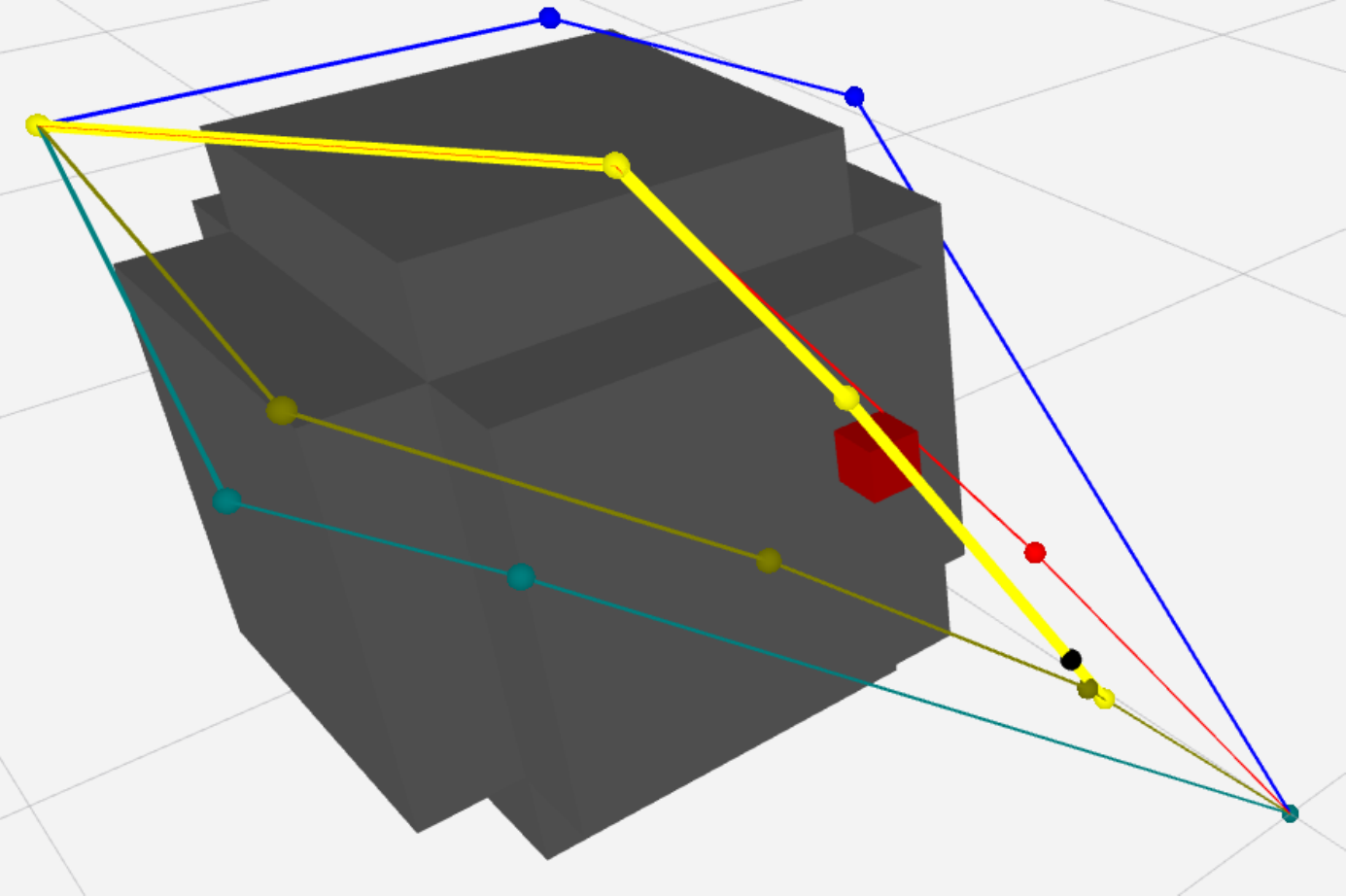}}\hfil
    \subcaptionbox{The pink path is the path the robot actually has crossed, result of re-plans and optimizations during robot motion. \label{subfig:opt}}
    {\includegraphics[width=0.4\linewidth]{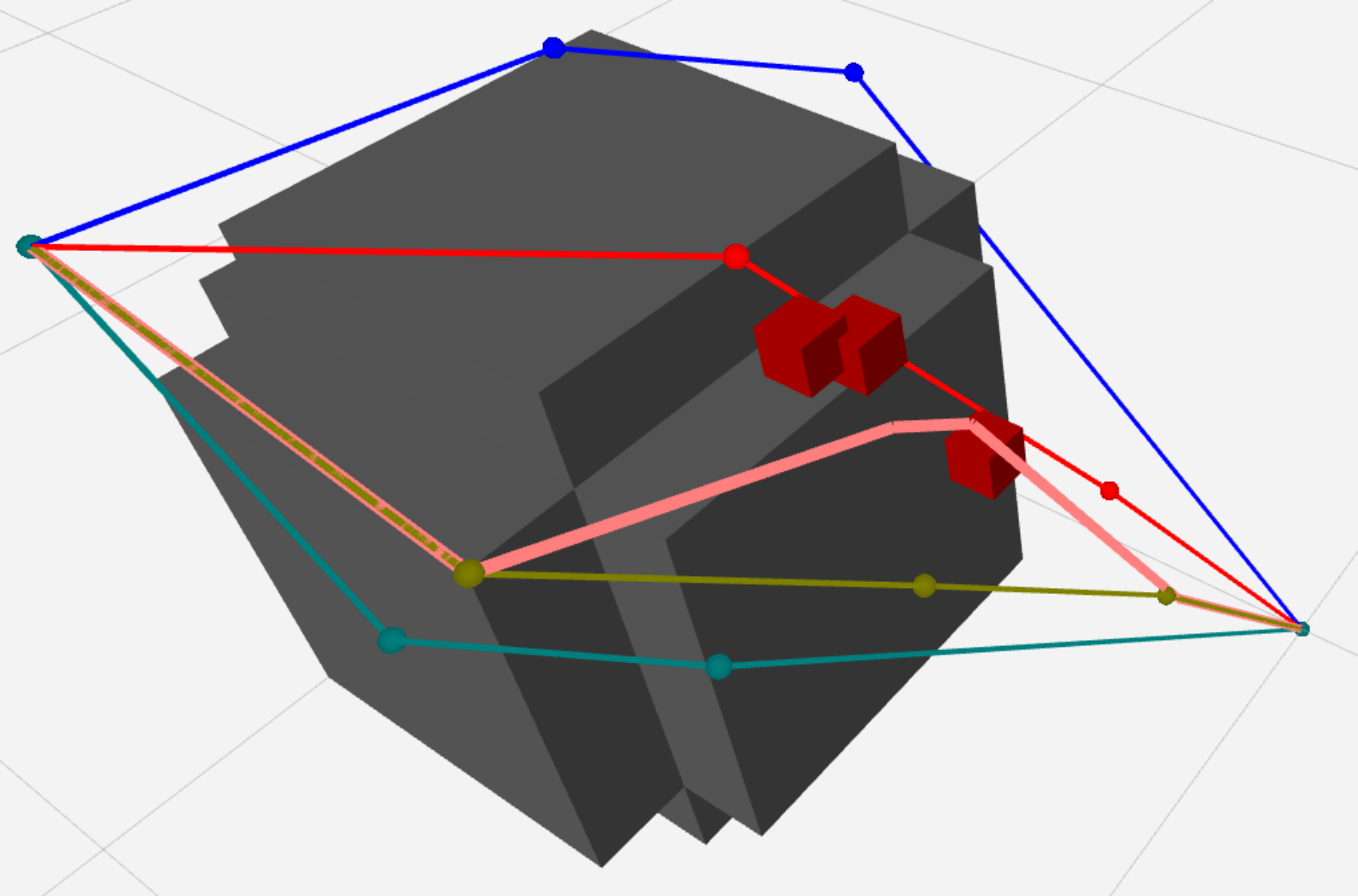}}
\caption{Example of a test with the simple 3D cell}
\label{fig:test}
\end{figure*}

Figure \ref{fig:test} shows the strategy working during a test in the simple cell. Four paths are computed at the start, the green one is the current path $\sigma_i$, the red, blue and light blue ones are the other available paths $\sigma_j \in S$ and the black sphere is the robot moving on the path. In Figure  \ref{subfig:init} the re-planner has optimized the current path finding the yellow one $\sigma_{\mathrm{RP}}$, which has a lower cost. 
Then a new obstacle obstructs it, so the yellow path becomes infeasible (Figure \ref{subfig:obstacle}). So, the algorithm finds a new free path that avoids it (Figure \ref{subfig:avoidance}). This path will be optimized in the next iterations. Finally, the path that the robot actually has crossed during the test is shown in pink (Figure \ref{subfig:opt}).

The solutions found are evaluated in terms of the time the algorithm has taken to find them and in terms of the relative variation of the length of the solution, $\left\Vert \sigma_{\mathrm{RP}}\right\Vert$, compared to the length of the path the robot was following before finding it, $\left\Vert \sigma_{\mathrm{cur}}\right\Vert$, defined as follows:
\begin{equation}
\Delta = 100 \cdot \frac{\left\Vert \sigma_{\mathrm{cur}}\right\Vert-\left\Vert \sigma_{\mathrm{RP}} \right\Vert}{\left\Vert \sigma_{\mathrm{cur}}\right\Vert}
\label{eq:quality-index}
\end{equation}

Tables \ref{tab:simple_cell} and \ref{tab:complex_cell} show the results for the 3D and the 6D scenarios, respectively. In particular, the mean and standard deviation of the path length variations \eqref{eq:quality-index} and of the re-planning times, and the number of re-plannings have been reported. 
Results are divided in two cases: when the re-planner aims to avoid an obstacle and when it optimizes the current path.
As expected, the time required to re-plan in the 6D scenario is larger than that required in the 3D scenario, but the re-planning algorithm has always found a solution before the available time expired. 
In case of obstacle avoidance, the path length tends to increase because  the algorithm tries to quickly find a first feasible solution, giving less importance to its optimization. Furthermore, when a new obstacle obstructs the current path, it is reasonable that the new solution is longer, since it has to circumvent the new obstacle.
In the 3D scenario the path improvement after optimization is small (mean($\Delta$)=$2$\%); this is due to the fact that since the robot has a very simple kinematic structure, the initial paths are close to be optimal.
On the contrary, in the 6D scenario, a significant average reduction in the length of the paths can be noted for the optimization case (mean($\Delta$)=$16.9$\%).
However, for the same reason, the solutions found every times the obstacle obstructs the current path suffers from a big increment of the path length (mean($\Delta$)=$-161$\%). 
Nonetheless, the \emph{re-planning thread} will keep on improving such solution during the execution.

\begin{table}[t]
\caption{Results of the 3D scenario}
\label{tab:simple_cell}
\centering
\begin{tabular}{|l|l|l|}
\hline
Simple cell          & Obstacle avoidance        &    Path optimization  \\
\hline
mean($\Delta$) (\%)            & -8.41     & 2.00          \\
std. deviation($\Delta$) (\%)        & 24.7    & 6.08           \\
mean(time) (ms)          & 0.0146     & 0.00509           \\
std. deviation(time) (ms)           & 0.0117     & 0.00478           \\
numer of re-plans          & 90     & 148           \\
\hline
\end{tabular}
\end{table}

\begin{table}[t]
\caption{Results of the 6D scenario}
\label{tab:complex_cell}
\centering
\begin{tabular}{|l|l|l|}
\hline
Complex cell          & Obstacle avoidance        &    Path optimization  \\
\hline
mean($\Delta$) (\%)            &-161      & 16.9          \\
std. deviation($\Delta$) (\%)        &141     & 23.4           \\
mean(time) (ms)          &0.0438      &0.0280            \\
std. deviation(time) (ms)           &0.0112      &0.0325            \\
number of re-plans           & 90     &251            \\
\hline
\end{tabular}
\end{table}

\section{Conclusions}
\label{sec:conclusions}

We have proposed an anytime path re-planning framework with double functionality to continuously optimize the current path and to find a new feasible path when a new obstacle obstructs it. The strategy exploits a set of pre-computed paths and efficiently tries to connect to them to find or improve the current solution. Numerical results show the effectiveness of the strategy in different scenarios.
Future works will focus on ensuring that not only the path is collision-free, but it is also robust with respect to tracking errors introduced to fulfill velocity and acceleration constraints.

\bibliographystyle{IEEEtran}
\bibliography{reference_stiima}

\begin{thebibliography}{10}
\providecommand{\url}[1]{#1}
\csname url@samestyle\endcsname
\providecommand{\newblock}{\relax}
\providecommand{\bibinfo}[2]{#2}
\providecommand{\BIBentrySTDinterwordspacing}{\spaceskip=0pt\relax}
\providecommand{\BIBentryALTinterwordstretchfactor}{4}
\providecommand{\BIBentryALTinterwordspacing}{\spaceskip=\fontdimen2\font plus
\BIBentryALTinterwordstretchfactor\fontdimen3\font minus
  \fontdimen4\font\relax}
\providecommand{\BIBforeignlanguage}[2]{{%
\expandafter\ifx\csname l@#1\endcsname\relax
\typeout{** WARNING: IEEEtran.bst: No hyphenation pattern has been}%
\typeout{** loaded for the language `#1'. Using the pattern for}%
\typeout{** the default language instead.}%
\else
\language=\csname l@#1\endcsname
\fi
#2}}
\providecommand{\BIBdecl}{\relax}
\BIBdecl

\bibitem{A_star}
P.~E. Hart, N.~J. Nilsson, and B.~Raphael, ``A formal basis for the heuristic
  determination of minimum cost paths,'' \emph{IEEE transactions on Systems
  Science and Cybernetics}, vol.~4, no.~2, pp. 100--107, 1968.

\bibitem{RRT}
S.~LaValle, ``Rapidly-exploring random trees: a new tool for path planning,''
  \emph{The annual research report}, 1998.

\bibitem{karaman:RRT*}
S.~Karaman and E.~Frazzoli, ``Sampling-based algorithms for optimal motion
  planning,'' \emph{The International Journal of Robotics Research}, vol.~30,
  no.~7, pp. 846--894, 2011.

\bibitem{Gammell2020}
J.~D. Gammell and M.~P. Strub, ``Asymptotically optimal sampling-based motion
  planning methods,'' \emph{Annual Review of Control, Robotics, and Autonomous
  Systems}, vol.~4, no.~1, pp. 19.1--19.24, 2021.

\bibitem{Gammel:InformedRRT}
J.~D. Gammell, T.~D. Barfoot, and S.~S. Srinivasa, ``Informed sampling for
  asymptotically optimal path planning,'' \emph{IEEE Transactions on Robotics},
  vol.~34, no.~4, pp. 966--984, 2018.

\bibitem{Gammell:BIT}
J.~D. {Gammell}, T.~D. {Barfoot}, and S.~S. {Srinivasa}, ``Batch informed trees
  ({BIT*}): Informed asymptotically optimal anytime search,''
  \emph{International Journal of Robotics Research}, vol.~39, no.~5, pp.
  543--567, 2020.

\bibitem{Aine2016}
S.~Aine and M.~Likhachev, ``Truncated incremental search,'' \emph{Artificial
  Intelligence}, vol. 234, pp. 49--77, 2016.

\bibitem{Magrini:coexistance-interaction}
E.~Magrini, F.~Ferraguti, A.~Ronga, F.~Pini, A.~Luca, and F.~Leali,
  ``Human-robot coexistence and interaction in open industrial cells,''
  \emph{Robotics and Computer-Integrated Manufacturing}, vol.~61, p. 101846,
  2020.

\bibitem{Zanchettin:safety_HRC}
A.~M. Zanchettin, N.~M. Ceriani, P.~Rocco, H.~Ding, and B.~Matthias, ``Safety
  in human-robot collaborative manufacturing environments: Metrics and
  control,'' \emph{IEEE Transactions on Automation Science and Engineering},
  vol.~13, no.~2, pp. 882--893, 2016.

\bibitem{Faroni_UR2020}
M.~Faroni, R.~Pagani, and G.~Legnani, ``Real-time trajectory scaling for robot
  manipulators,'' in \emph{Proceedings of the International Conference on
  Ubiquitous Robots}, Kyoto (Japan), 2020.

\bibitem{Faroni_ETFA2019}
M.~Faroni, M.~Beschi, and N.~Pedrocchi, ``An {MPC} framework for online motion
  planning in human-robot collaborative tasks,'' in \emph{Proceedings of the
  IEEE Int. Conf. on Emerging Tech. and Factory Automation}, Zaragoza (Spain),
  2019.

\bibitem{RRF}
T.-Y. Li and Y.-C. Shie, ``An incremental learning approach to motion planning
  with roadmap management,'' in \emph{Journal of Information Science and
  Engineering}, vol.~23, 2002, pp. 3411 -- 3416.

\bibitem{DRRT}
D.~Ferguson, N.~Kalra, and A.~Stentz, ``{Replanning with {RRTs}},'' in
  \emph{Proceedings of the IEEE International Conference on Robotics and
  Automation}, 2006, pp. 1243--1248.

\bibitem{Connell:DRRT*}
D.~Connell and H.~La, ``Dynamic path planning and replanning for mobile robots
  using {RRT},'' in \emph{IEEE International Conference on Systems, Man, and
  Cybernetics}, 2017, pp. 1429--1434.

\bibitem{MP-RRT}
M.~Zucker, J.~Kuffner, and M.~Branicky, ``{Multipartite {RRTs} for rapid
  replanning in dynamic environments},'' \emph{Proceedings of the IEEE
  International Conference on Robotics and Automation}, pp. 1603--1609, 2007.

\bibitem{ERRT}
J.~Bruce and M.~M. Veloso, ``{Real-time randomized path planning for robot
  navigation},'' \emph{Lecture Notes in Artificial Intelligence (Subseries of
  Lecture Notes in Computer Science)}, vol. 2752, pp. 288--295, 2003.

\bibitem{RRTX}
M.~Otte and E.~Frazzoli, ``{RRTx: Real-time motion planning/replanning for
  environments with unpredictable obstacles},'' \emph{The International Journal
  of Robotics Research}, vol.~35, no.~7, pp. 797--822, 2016.

\bibitem{HL-RRT*}
Y.~Chen, Z.~He, and S.~Li, ``{Horizon-based lazy optimal {RRT} for fast,
  efficient replanning in dynamic environment},'' \emph{Autonomous Robots},
  vol.~43, no.~8, pp. 2271--2292, 2019.

\bibitem{Zhang2021}
Z.~Zhang, B.~Qiao, W.~Zhao, and X.~Chen, ``{A Predictive Path Planning
  Algorithm for Mobile Robot in Dynamic Environments Based on Rapidly Exploring
  Random Tree},'' \emph{Arabian Journal for Science and Engineering}, 2021.

\bibitem{van-den-Berg:anytime}
J.~van~den Berg, D.~Ferguson, and J.~Kuffner, ``Anytime path planning and
  replanning in dynamic environments.'' in \emph{Proceedings of the IEEE
  International Conference on Robotics and Automation}, 2006, pp. 2366--2371.

\bibitem{elastic-strips}
O.~Brock and O.~Khatib, ``Elastic strips: A framework for motion generation in
  human environments,'' \emph{International Journal of Robotic Research},
  vol.~21, pp. 1031--1052, 2002.

\bibitem{Lavalle:RRTconnect}
J.~J. Kuffner and S.~M. LaValle, ``{RRT-Connect}: An efficient approach to
  single-query path planning,'' in \emph{Proceedings of the IEEE International
  Conference on Robotics and Automation}, vol.~2, San Francisco (USA), 2000,
  pp. 995--1001.

\end{thebibliography}

\end{document}